\documentclass[11pt,a4paper]{article}
\usepackage[hyperref]{emnlp2020}
\usepackage{times}
\usepackage{latexsym}

\usepackage{microtype}

\usepackage{graphicx}
\usepackage{amsfonts,amssymb,amsmath}
\usepackage{mathtools}
\usepackage{tabu}
\usepackage{multirow}
\usepackage{algorithm}
\usepackage{algorithmic}
\usepackage{xspace}
\usepackage{booktabs}

\aclfinalcopy 
\def\aclpaperid{3633} 

\title{The Volctrans Machine Translation System for WMT20}

\author{Liwei Wu$^1$, Xiao Pan$^1$, Zehui Lin$^2$\thanks{\ \ Intern at ByteDance}, Yaoming Zhu$^3$\thanks{\ \ Intern at ByteDance}, Mingxuan Wang$^1$, Lei Li$^1$ \\
  $^1$ByteDance AI Lab, Beijing, China \\
  \textit{\{wuliwei.000, panxiao.94, wangmingxuan.89, lilei.lab\}@bytedance.com} \\
  $^2$Fudan University, Shanghai China \\
  \textit{linzh18@fudan.edu.cn} \\
  $^3$Shanghai Jiao Tong University, Shanghai China \\
  \textit{ymZhu@apex.sjtu.edu.cn} \\
  }

\date{}

\begin{document}
\maketitle

\begin{abstract}

This paper describes our VolcTrans system on WMT20 shared news translation task. We participated in 14 translation directions. 
Our basic systems are based on Transformer \cite{vaswani2017attention}, with several variants (wider or deeper Transformers, dynamic convolutions). The final system includes text pre-process, data selection,  synthetic data generation, advanced model ensemble, and multilingual pre-training.  

\end{abstract}
\section{Introduction}
\label{sec:intro}

We participated in the WMT2020 shared news translation task in 14 directions: English$\leftrightarrow$Chinese, English$\leftrightarrow$German, French$\leftrightarrow$German, English$\leftrightarrow$Polish,
English$\leftrightarrow$Tamil,English$\leftrightarrow$Pashto,English$\leftrightarrow$Khmer, covering language pairs from high to low resources.  
In this year’s translation task, we  mainly focus on exploiting self-supervised and unsupervised methods for NMT to make full use of the monolingual data~\cite{lin2020pre,yang2019towards}. 

We aims at building a general training framework which can be well applied to different translation directions. 
Our models are mainly based on the Transformer \cite{vaswani2017attention}. Techniques used in the submitted systems include iterative back-translation, knowledge distillation. We also employed several tricks to improve in-domain BLEU scores, typically in-domain transfer learning. We also experimented with a multilingual pre-training technique which we proposed recently~\cite{lin2020pre}.

\section{Baseline Models}
\label{sec:baseline model}
We apply two different NMT skeletons for the shared news translation as our baseline systems. We use the implementations in Fairseq\cite{ott2019fairseq}. All models are trained with Adam optimizer \cite{kingma2014adam}. We use the ``inverse sqrt lr" scheduler with 4000 warm-up steps and set the max learning rate to 5e-4. The betas are (0.9, 0.98). During training, the batches are made of similar length sequences, so we avoid extreme cases where most sequences in the batch are short and we are required to add lots of pad tokens to each of them because one sequence of the same batch is very long. We limit the batch size to 8192 tokens per GPU, to avoid running out of GPU memory. Meanwhile, to achieve a larger batch size to improve the performance\cite{ott2018scalenmt}, we set the parameter ``update frequency" to 8, and train the model on 8 GPUs, resulting in an actual batch token size = $8192 \times 8 \times 8$. During training, we employ label smoothing of 0.1 and set dropout rate \cite{hinton2012improving} to 0.2.

\subsection{Transformer}
Following \citet{sun2019baidu,wang2018tencent}, we use  different architectures for Transformer\cite{vaswani2017attention} to increase the model diversity and potentially get a better ensemble model.

\begin{itemize}
	\item Transformer 15e6d: According to \citet{sun2019baidu}, a transformer with larger encoder layer number can learn better representation of source sentence and get better BLEU scores. We increase the number of encoder layers from 6 to 15 layers in the transformer big architecture which is the same as the Deeper Transformer in \citet{sun2019baidu}.
	\item Transformer Mid 25e6d and Transformer Mid 50e6d: To get much better BLEU scores, we further increase the encoder layer number from 6 to 25 (Transformer Mid 25e6d) and 50 (Transformer Mid 50e6d) for the transformer big architecture. However, the model is too large and can not be trained with GPU, so we decrease the feed forward size from 4096 to 3072 and the embedding size from 1024 to 768.
	\item Transformer 15000ffn. According to \citet{sun2019baidu}, the performance of the Transformer model is largely dependent on the dimensions of feed forward network. We use the same architecture as Bigger Transformer in \citet{sun2019baidu} which increases the  feed forward size from 4096 to 15000, the attention dropout from 0.1 to 0.3 and the relu dropout from 0.1 to 0.3. The number of encoder and decoder layers remains 6.
	\item Transformer 128hdim and Transformer 256hdim. \citet{bhojanapalli2020low} shows that a transformer model with larger attention dimensions can also get better BLEU score. We increase the head dimension from 64 to 128 (Transformer 128hdim) and 256 (Transformer 256hdim). The number of encoder and decoder layers remains 6.
	\item DLCL 25layers. \citet{DBLP:conf/wmt/LiLXLLLWZXWFCLL19} proposes a transformer variant call DLCL and shows that this architecture can make deep transformer get higher BLEU.
\end{itemize}

\subsection{Dynamic Convolution}
We also apply dynamic convolution \cite{wu2019pay} architectures.

\begin{itemize}
    \item Dynamic Convolution 7e6d: The dynamic convolution model with 7 encoder layers and 6 decoder layers which is the same architecture proposed in \citet{wu2019pay}.
    \item Dynamic Convolution 25e6d: We increase the encoder layer number from 6 to 25. For layers above 7, we set the kernel size to 31.
\end{itemize}

\section{Experiment Techniques}
\label{section:experiment_techniques}
\subsection{Parallel Data Up-sampling}
\label{sub_section:sample_data}
According to the experiments, data diversity matters for the whole system. Apart from splitting the monolingual data into several disjoint parts, we sampled the parallel data so that each model has different deviations on the parallel data. We tested bagging sampling (sample with replacement) and up-sampling(sample with replacement under the premise of using all data), experimental results show that when the amount of parallel data is inadequate with respect to the amount of model parameters (such as French$\leftrightarrow$German, English$\leftrightarrow$Polish, etc.), the bagging sampling method reduces the performance of the model; while when the amount of parallel data is abundant(such as English $\leftrightarrow$ German), the bagging sampling method has no significant effects on the performance. On the contrary, the data up-sampling method never degrades the performance of the model.

\paragraph{mRASP: Multilingual Pre-training}
\label{subsection:PNMT}
We employed a pre-training method mRASP, which pre-trains a universal  multilingual  neural  machine  translation  model and fine-tune it on specific language directions. Basically, we pre-train a model using the provided parallel data on WMT2020 of English$\leftrightarrow$Khmer, English$\leftrightarrow$Inuktitut, French$\leftrightarrow$German, English$\leftrightarrow$Polish, English$\leftrightarrow$Pashto, English$\leftrightarrow$Tamil, on a shared vocabulary learned from the above parallel data plus provided monolingual data of all related languages. We learn a BPE sub-word vocabulary with 6000 merge operations. We up-sample the data from lower resource language data to balance data amount and only keep tokens that occur more than 10 times. Finally, we obtain a joint vocabulary of about 28000 tokens.

We fine-tuned the pre-trained model, for low-resource directions: Pashto$\rightarrow$English, English$\leftrightarrow$Khmer and English$\leftrightarrow$Tamil. The baseline model initialized by this method performs better than the randomly initialized baseline model by a large margin. We pre-trained three mRASP models using the same training data: Transformer big, Transformer 15000ffn and Dynamic Convolution. We report in Table~\ref{tab:pretrain} the best score in each setting and direction, and find that mRASP significantly outperforms the baseline.

\begin{table*}[ht]
\begin{center}

\begin{tabular}{l|c|c|c|c|c}
\toprule
 Testset & 
Ps$\rightarrow$En & En$\rightarrow$Km & Km$\rightarrow$En & En$\rightarrow$Ta & Ta$\rightarrow$En
\tabularnewline 

\midrule
 Random &
10.2 & 39.3 & 12.7 & 7.4 & 14.0
\tabularnewline

w/ mRASP & 
13.8 & 42.8 & 14.4 & 9.2 & 17.9
\tabularnewline

\bottomrule
\end{tabular}

\end{center}
\caption{Comparison between randomly initialized baseline model and model initialized from mRASP model}
\label{tab:pretrain}
\end{table*}

\subsection{Tag Back-Translation}
Recently, back-translation \cite{edunov2018understanding} is a standard method to improve the translation quality by leveraging the large scale monolingual data. Starting from WMT19, the source of the test set is the natural text and the  of the test set is the translationese text. We find the tag back-translation \cite{caswell2019tagged} method can achieve better BLEU compared with previous methods proposed in \citet{edunov2018understanding}. 

\begin{figure}[!t]
    \centering
    \includegraphics[width=.5\textwidth]{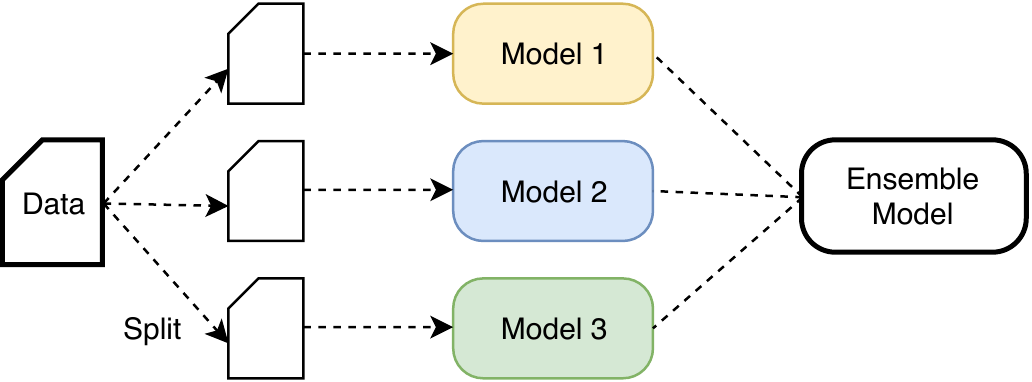}
    \caption{Data Diversity Matters for Final System}
    \label{fig:tag_bt}
\end{figure}

To improve the data diversity among single models before model ensemble, we generated the back-translated data from different monolingual data using different baseline models, as illustrated in Figure~\ref{fig:tag_bt}. For high resource data (English, Polish, etc.), we divided monolingual data into several parts, each containing 10M sentences. However, for low resource data (Pashto), due to the lack of monolingual data, we use all monolingual data for all back-translation tasks. 

\subsection{Iterative Joint Training}
\citet{zhang2018joint} proposed an iterative joint training method for better usage of monolingual data from source side and target side. In each iteration, the S2T(source to target) model generates a S2T(target to source) synthetic data from the source side monolingual data and the T2S model generates a T2S synthetic data from the target side monolingual data. Then, the S2T and T2S model are trained with the new T2S and S2T synthetic data to improve the both models performance. In the next iteration, the S2T and T2S model can generate synthetic data with better quality and their performance can be improved further. We jointly trained the S2T and T2S model until they converge. Experiment results on English$\leftrightarrow$Polish shown in Table~\ref{tab:en_pl_joint_train}

\begin{table}[ht]
\begin{center}
\scalebox{0.85}{
\begin{tabular}{l|c|c}
\toprule
 Direction & 
\multicolumn{1}{c|}{{En$\rightarrow$Pl}} &
\multicolumn{1}{c}{{Pl$\rightarrow$En}} 
\tabularnewline
\midrule
 Testset &
news20 dev & 
news20 dev \tabularnewline

\midrule
 Baseline &
24.8 & 29.7
\tabularnewline

 Iter 1 &
27.5 & 32.6
\tabularnewline

 Iter 2 &
27.8 & 32.7
\tabularnewline

 Iter 3 &
28.2 & 33.3
\tabularnewline

\bottomrule

\end{tabular}
}
\end{center}
\caption{Iterative Joint Training for English$\leftrightarrow$Polish}
\label{tab:en_pl_joint_train}
\end{table}

\subsection{Knowledge Distillation}
Recently, knowledge distillation has been widely used to improve the performance of models \cite{sun2019baidu, DBLP:conf/wmt/LiLXLLLWZXWFCLL19}. In our knowledge distillation method, student model is trained to fit the output of teacher models. Concretely, we translate the source side monolingual data with an ensemble teacher and a right-to-left(R2L) \cite{DBLP:conf/naacl/LiuUFS16} model teacher.

\begin{itemize}
    \item Ensemble Model. We divided single models in the last joint training iteration into k groups (k=3 in our experiments, resulting in 3 models in each group) and ensemble models in one group to as the teacher model.
    \item R2L Model. We trained one R2L model for each ensemble group using the same data as anyone model in this group from the last iteration. 
\end{itemize}

We then use pseudo parallel data from ensemble model as well as from R2L model to train the student model, without employing parallel data.

\subsection{Advanced Tricks}
\paragraph{Top-k Checkpoint Average}
Different from the conventional checkpoint average approach, which is to average continuous K checkpoints, we average K checkpoints which have the highest BLEU scores on the valid set, and find that this strategy usually leads to significant BLEU improvements over single checkpoints.

\begin{figure}[!t]
    \centering
    \includegraphics[width=.5\textwidth]{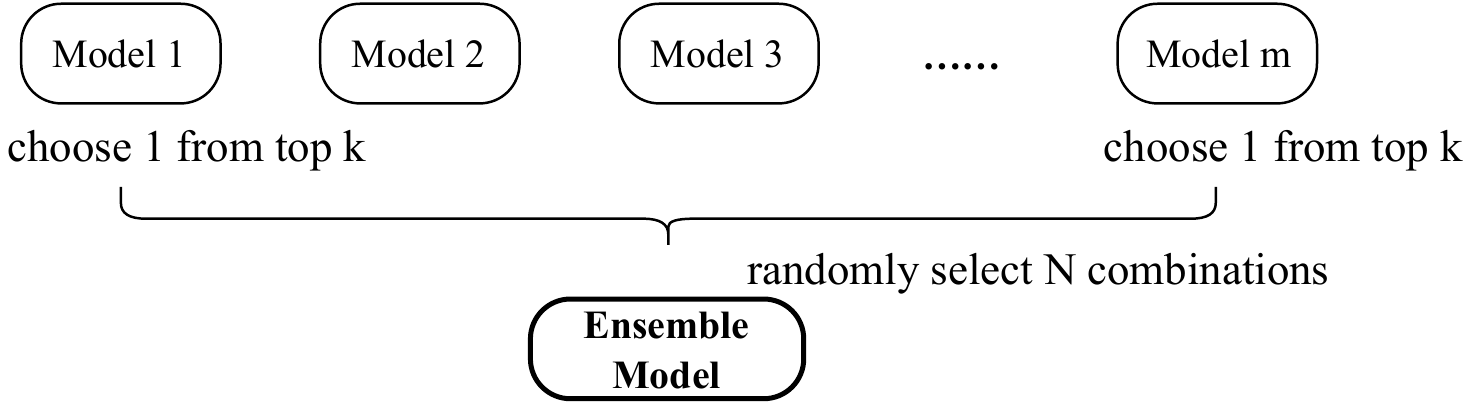}
    \caption{Illustration of Random Ensemble}
    \label{fig:random_ensemble}
\end{figure}

\paragraph{Random Ensemble}
We adopt a simple yet effective strategy in model ensemble. Rather than select the best checkpoint from each model (a.k.a. greedy search), we enlarge the search space: choose one checkpoint from top-k checkpoints from each model, and randomly select N combinations from the entire search space, see Figure~\ref{fig:random_ensemble}.

\paragraph{In domain Fine-tuning}
There exists a domain mismatch between the obtained system trained with provided parallel or monolingual training data and the target test set. In order to alleviate this mismatch, and to improve the translation performance in the domain of the target test set, we fine-tune the best single models with development sets for 1-2 epochs.
\section{Settings and Results}
For all news tasks, all ParaCrawl corpus is cleaned by the script proposed by \citet{xia2019microsoft}. We trained the baseline using the sampling method described in Section~\ref{sub_section:sample_data} with different architectures showed in Section~\ref{sec:baseline model}. For the low resource language pair (English$\leftrightarrow$Pashto), as showed in Section~\ref{subsection:PNMT}, we pretrained three multilingual models with different model architectures (DLCL 25layers, Transformer 15000ffn and Dynamic Convolution 25e6d) on all parallel data available in WMT20 except the English$\leftrightarrow$Chinese to avoid a large dictionary\footnote{Large dictionary leads to large parameter size in embedding} and fine-tuned the pre-trained models on the their own parallel data with different data sampling strategies to get 9 baseline models\footnote{We randomly combine sampling strategies and model architectures to get 9 baseline models for each direction. The performances of the 9 baselines is not the point, what we want is the model diversity among single models}. Then we applied the tag back-translation, joint training, knowledge distillation and random ensemble methods as described in Section~\ref{section:experiment_techniques} to get the final translation system. All BLEU scores were reported with SacreBLEU\cite{post-2018-call}. 

\begin{table}[ht]
\begin{center}
\scalebox{0.85}{
\begin{tabular}{l|c}
\toprule
 Direction & 
\multicolumn{1}{c}{{En$\rightarrow$Zh}} 
\tabularnewline
\midrule
 Testset &
wmt19 \tabularnewline 

\midrule
 Baseline &
38.5 
\tabularnewline

 iterative BT &
38.9  
\tabularnewline

 Ensemble KD &
41.5 
\tabularnewline

 Ensemble System &
42.0 
\tabularnewline

 \midrule
\textbf{BLEU on WMT20} & \multirow{2}{*}{44.9}   \tabularnewline
 \textbf{testset submission}
\tabularnewline

\bottomrule

\end{tabular}
}
\end{center}
\caption{Results of English$\rightarrow$Chinese by sacreBLEU}
\label{tab:en_zh}
\end{table}

\subsection{Chinese$\rightarrow$English}
\paragraph{Final Submission} We submitted our VolcTrans online system (unconstrained). The final submission achieves 36.6 BLEU. You can get access to VolcTrans online system on \url{http://translate.volcengine.cn/}.

\subsection{English$\rightarrow$Chinese}
For English$\rightarrow$Chinese, we train English$\leftrightarrow$Chinese jointly. We use all parallel data available: News Commentary v15, Wiki Titles v2, UN Parallel Corpus V1.0, CCMT Corpus and WikiMatrix. After data filtering, XM parallel data remained. We use MosesTokenizer for English and Jieba for Chinese. After the pre-processing, separate BPE vocabulary is learned with 32000 merge operations for both English and Chinese on the parallel data. We sample parallel data of ratio 100\%, 110\% and 120\% with replacement from all parallel data. Then we train 3 baselines with Transformer Mid 25e6d, Transformer Mid 50e6d and Dynamic Convolution 25e6d architectures respectively, resulting in 9 baseline models. We employ Newscrawl data as monolingual data for English. The total amount of monolingual data is 90M, containing all Newscrawl 2019 data and others sampled from Newscrawl 2014 to 2018. For Chinese, we employ Newscrawl data, CCMT data and LDC data. We split the Chinese into 3 parts, each contains 8M sentences. For iterative back translation stage and ensemble knowledge distillation stage, each model is combined with different English monolingual data part. Since there are only 3 parts of Chinese monolingual data, we use each part for 3 times at each stage. At the ensemble knowledge distillation stage, we also employ disjoint monolingual data as the distilling data. The detailed results of our system is reported in Table \ref{tab:en_zh}. 

\paragraph{Final Submission} We submitted the ensemble system of the 9 single models after ensemble knowledge distillation stage. The final submission achieves 44.9 BLEU.

\subsection{English$\leftrightarrow$German}

For English$\leftrightarrow$German, we train both directions jointly. We use all parallel data available: Europarl v10, ParaCrawl v5.1, Common Crawl corpus, News Commentary v15, Wiki Titles v2, Tilde Rapid corpus and WikiMatrix corpus. After data filtering, 28M parallel data remained. We use MosesTokenizer for both English and German. After the pre-processing, a joint BPE vocabulary is learned with 6000 merge operations on the parallel data. We sample parallel data of ratio 80\%, 90\% and 100\% with replacement from all parallel data. Then we train 3 baselines with Transformer Mid 25e6d, Transformer Mid 50e6d and Dynamic Convolution 25e6d architectures respectively, resulting in 9 baseline models. We only employ Newscrawl data as monolingual data for both German and English. The total amount of monolingual data is 90M, containing all Newscrawl 2019 data and others sampled from Newscrawl 2014 to 2018. The 90M data was divided into 9 disjoint parts, each containing 10M sentences, to jointly train 9 systems separately. At the ensemble knowledge distillation stage, we also employ disjoint monolingual data as the distilling data. The detailed results of our system is reported in Table~\ref{tab:en_de}. 
\begin{table}[ht]
\centering
\scalebox{0.8}{
\begin{tabular}{l|cc|cc}
\toprule
 Direction & \multicolumn{2}{c|}{En$\rightarrow$De} & \multicolumn{2}{c}{De$\rightarrow$En} \tabularnewline
\midrule
 Testset & news18 & news19 & news18 & news19 \tabularnewline
\midrule
 Baseline & 47.1 & 42.2 & 45.7 & 41.6  \tabularnewline
 iterative BT & 48.6 & 42.6 & 48.1 & 42.2  \tabularnewline
 KD & 49.7 & 44.3 & 48.4 & 43.3 \tabularnewline
 
 \midrule
 Ensemble System & \textbf{52.2} & \textbf{46.1} & \textbf{49.1} & \textbf{43.8} \tabularnewline

 \midrule
 \textbf{BLEU on WMT20} & \multicolumn{2}{c|}{\multirow{2}{*}{38.2}} & \multicolumn{2}{c}{\multirow{2}{*}{43.5}} \tabularnewline
 \textbf{testset submission} & &
 
\\
 
\bottomrule
\end{tabular}
}
\caption{Results of English$\leftrightarrow$German by sacreBLEU}
\label{tab:en_de}
\end{table}

\paragraph{Fine-tune}
In this step we use the development sets to handle the domain mismatch problem in WMT. For English$\rightarrow$German direction, we fine-tune some of the best single models on news2018 for 1-2 epochs, and then get the final ensemble model from models with fine-tune and models without fine-tune.

\paragraph{Final Submission}
For either direction, the final submission is an ensemble system from single models with highest BLEU scores on development sets.
For the final English$\rightarrow$German submission, we replaced the English quote with the German quote. The final submissions on Test20 data achieve 38.2 BLEU for English$\rightarrow$German direction and 43.5 BLEU for German$\rightarrow$English direction.

\subsection{French$\leftrightarrow$German}
For French$\leftrightarrow$German, we train both directions jointly. The overall parallel data contains 13M sentences available including: Europarl v10, ParaCrawl v5.1, Common Crawl corpus, News Commentary v15, Wiki Titles v2 and WikiMatrix corpus. We train 9 baseline models, each with different architectures (Transformer 15e6d * 2, Transformer Mid 25e6d, Transformer Mid 50e6d, Transformer 15000ffn, Transformer 128hdim, Transformer 256hdim and Dynamic Convolution 25e6d) and each group of three models is combined with 3 different sampling strategies (no sample, up sample 120\%, up sample 140\%)\footnote{There is little difference among the models with different sampling ratios, what we are concerned about is the data diversity caused by different sampling ratios.}, resulting in 9 single models for each direction. We only employ Newscrawl data as monolingual data for both German and French. The monolingual data contains 90M sentences, including all Newscrawl 2019 data and others are sampled from Newscrawl 2014 to 2018. The data of 90M pairs is divided into 9 disjoint parts, each containing 10M sentences to jointly train 9 systems separately. The detailed experiment results are shown in Table~\ref{tab:fr_de}. 

\paragraph{Final Submission}
 For either direction, the final submission is an ensemble system of all 9 models obtained after the knowledge distillation stage. For the final German$\rightarrow$French submission, we replaced the English quote with the French quote. The final submissions on Test20 data achieve 35.7 BLEU for French$\rightarrow$German direction and 35.3 BLEU for German$\rightarrow$French direction.
\begin{table}[ht]
\begin{center}
\scalebox{0.85}{
\begin{tabular}{l|c|c}
\toprule
 Direction & 
\multicolumn{1}{c|}{{Fr$\rightarrow$De}} &
\multicolumn{1}{c}{{De$\rightarrow$Fr}} 
\tabularnewline
\midrule
 Testset &
news19 & 
news19 \tabularnewline 
\midrule
 Baseline &
26.7 & 31.3
\tabularnewline

 iterative BT &
32.4 & 35.6
\tabularnewline

 KD &
32.7 & 36.8
\tabularnewline

\midrule
 Ensemble System & \textbf{33.9} & \textbf{38.0} \tabularnewline

\midrule
\textbf{BLEU on WMT20} & \multirow{2}{*}{35.7} & \multirow{2}{*}{35.3} \tabularnewline
 \textbf{testset submission} &
  &  
\tabularnewline

\bottomrule

\end{tabular}
}
\end{center}
\caption{Results of French$\leftrightarrow$German by sacreBLEU}
\label{tab:fr_de}
\end{table}

\subsection{English$\leftrightarrow$Polish}
\begin{table}[ht]
\begin{center}
\scalebox{0.85}{
\begin{tabular}{l|c|c}
\toprule
 Direction & 
\multicolumn{1}{c|}{{En$\rightarrow$Pl}} &
\multicolumn{1}{c}{{Pl$\rightarrow$En}} 
\tabularnewline
\midrule
 Testset &
news20 dev & 
news20 dev \tabularnewline 

\midrule
 Baseline &
24.8 & 29.7
\tabularnewline

iterative BT &
27.8 & 32.7
\tabularnewline

KD &
28.2 & 33.3
\tabularnewline

\midrule
Ensemble System &
 \textbf{28.7} & \textbf{34.0}
\tabularnewline

\midrule
\textbf{BLEU on WMT20} & \multirow{2}{*}{26.1} & \multirow{2}{*}{34.4} \tabularnewline
 \textbf{testset submission} &
  & 
\tabularnewline

\bottomrule

\end{tabular}
}
\end{center}
\caption{Results of English$\leftrightarrow$Polish by sacreBLEU}
\label{tab:en_pl}
\end{table}

Our English$\leftrightarrow$Polish systems are based on Europarl v10, ParaCrawl v5.1, Wiki Titles v2, Tilde Rapid corpus and WikiMatrix corpus. All data add up to 8M sentences. We train 9 models with different architectures (Transformer 15e6d * 2, Transformer Mid 25e6d, Transformer Mid 50e6d, Transformer 15000ffn, Transformer 128hdim, Transformer 256hdim and Dynamic Convolution 25e6d) and each group of three models is combined with different sampling strategies (no sample, up sample 120\%, up sample 140\%) on both directions. We only used Newscrawl as English monolingual data. The English monolingual data contains 90M sentences, including all Newscrawl 2019 data and others sampled from Newscrawl 2014 to 2018. We divide the data into 9 disjoint parts. Since Polish Newscrawl corpus only contains 3M sentences, we additionally employ the Polish common crawl data. We sample 90M sentences from the Polish common crawl data which is then divided into 9 disjoint parts. Each part contains 10M common crawl sentences and 3M Newscrawl sentences. Then we apply the joint training and knowledge distillation as described in Section~\ref{section:experiment_techniques}. The detailed experiment results are shown in Table~\ref{tab:en_pl}. 

\paragraph{Final Submission}
Our final submission is an ensemble system consisting of all 9 models obtained after the ensemble knowledge distillation stage. The final submissions on Test20 data achieve 26.1 BLEU for English$\rightarrow$Polish direction and 34.4 BLEU for Polish$\rightarrow$English direction.

\subsection{English$\leftrightarrow$Pashto}
\begin{table}[ht]
\begin{center}
\scalebox{0.85}{
\begin{tabular}{l|c|c}
\toprule
 Direction & 
\multicolumn{1}{c|}{{En$\rightarrow$Ps}} &
\multicolumn{1}{c}{{Ps$\rightarrow$En}} 
\tabularnewline
\midrule
 Testset &
news20 dev & 
news20 dev \tabularnewline 

\midrule
 Baseline &
8.4 & 10.2
\tabularnewline

 +mRASP &
 - & 13.8
\tabularnewline

 iterative BT &
\textbf{9.6} & 16.4
\tabularnewline

 Ensemble System &
- & \textbf{18.0}
\tabularnewline

 \midrule
\textbf{BLEU on WMT20} & \multirow{2}{*}{10.6} & \multirow{2}{*}{20.0} \tabularnewline
 \textbf{testset submission} &
  & 
\tabularnewline

\bottomrule

\end{tabular}
}
\end{center}
\caption{Results of English$\leftrightarrow$Pashto by sacreBLEU}
\label{tab:en_ps}
\end{table}

For English$\leftrightarrow$Pashto, we used all parallel data containing 13M sentences available as follows: ParaCrawl v5.1, Wiki Titles v2 and the Khmer and Pashto parallel data. For Pashto$\rightarrow$English, we fine tune the three pre-trained models on all data with different sampling strategies. Each pre-trained model is fine tuned with three different sampling strategies and we get 9 models. For English$\rightarrow$Pashto, we find that the models fine-tuned from the pre-trained models have lower BLEU score than the baseline model trained from scratch, so we use the 9 baseline models which are trained with different architectures and sampling strategies. The English monolingual data has 90M sentences containing all Newscrawl 2019 data and others sampled from Newscrawl 2014 to 2018. We divide the data into 9 groups. The detailed experiment results are shown in Table~\ref{tab:en_ps}.

\paragraph{Final Submission}
For Pashto$\rightarrow$English, our final submission is an ensemble model consisting of all 9 models obtained after the ensemble knowledge distillation stage. For English$\rightarrow$Pashto, we find the ensemble model has lower BLEU score than the best single model, so we use the best single model as our final submission. The final submissions achieve 10.6 BLEU on wmt20 testset for English$\rightarrow$Pashto direction and 20.0 BLEU for Pashto$\rightarrow$English direction.

\subsection{English$\leftrightarrow$Tamil}
\begin{table}[ht]
\begin{center}
\scalebox{0.85}{
\begin{tabular}{l|c|c}
\toprule
 Direction & 
\multicolumn{1}{c|}{{En$\rightarrow$Ta}} &
\multicolumn{1}{c}{{Ta$\rightarrow$En}} 
\tabularnewline
\midrule
 Testset &
news20 dev & 
news20 dev \tabularnewline 

\midrule
 Baseline &
7.4 & 14.0
\tabularnewline

 +mRASP &
9.2 & 17.9
\tabularnewline

 iterative BT &
11.8 & 23.8
\tabularnewline

\midrule
\textbf{BLEU on WMT20} & \multirow{2}{*}{7.9} & \multirow{2}{*}{19.7} \tabularnewline
 \textbf{testset submission} 
\tabularnewline

\bottomrule

\end{tabular}
}
\end{center}
\caption{Results of English$\leftrightarrow$Tamil by sacreBLEU}
\label{tab:en_ta}
\end{table}

For English$\leftrightarrow$Tamil, we use all parallel data containing 533K sentences in total. We use all provided Tamil monolingual data. Following the procedure of English$\leftrightarrow$German, the English monolingual data contains 90M sentences, including all Newscrawl 2019 data and others sampled from Newscrawl 2014 to 2018. The 90M data was divided into 9 disjoint parts, each containing 10M sentences. For Tamil, we don't apply tokenizer and the raw text is directly split by BPE subword.
We fine-tune the three pre-trained models on all parallel data with two different sampling strategies: no sample and up sample 120\%, resulting in 6 models. We then conduct back translation for one iteration afterwards, each using one part of the English monolingual data for generating English$\rightarrow$Tamil pseudo data, and all Tamil monolingual data for generating Tamil$\rightarrow$English pseudo data. The detailed experiment results are shown in Table \ref{tab:en_ta}.

\paragraph{Final Submission}
For both English$\rightarrow$Tamil and Tamil$\rightarrow$English directions, our final submission is a single model. The final submissions achieve 7.9 BLEU for English$\rightarrow$Tamil and 19.7 BLEU for Tamil$\rightarrow$English.

\subsection{English$\leftrightarrow$Khmer}
\begin{table}[ht]
\begin{center}
\scalebox{0.85}{
\begin{tabular}{l|c|c}
\toprule
 Direction & 
\multicolumn{1}{c|}{{En$\rightarrow$Km}} &
\multicolumn{1}{c}{{Km$\rightarrow$En}} 
\tabularnewline
\midrule
 Testset &
news20 dev & 
news20 dev \tabularnewline 

\midrule
 Baseline &
39.3 & 12.7
\tabularnewline

 +mRASP &
42.8 & 14.4
\tabularnewline

 iterative BT &
46.5 & 16.9  
\tabularnewline

 Ensemble System &
- & 17.8 
\tabularnewline

 \midrule
\textbf{BLEU on WMT20} & \multirow{2}{*}{51.8} & \multirow{2}{*}{17.6}   \tabularnewline
 \textbf{testset submission} &
  & 
\tabularnewline

\bottomrule

\end{tabular}
}
\end{center}
\caption{Results of English$\leftrightarrow$Khmer by sacreBLEU}
\label{tab:en_km}
\end{table}

For English$\leftrightarrow$Khmer, we used all parallel data containing 4M sentences available as follows: ParaCrawl v5.1 and the Khmer and Pashto parallel data. For Khmer, we extract a dictionary from the Khmer and Pashto parallel data. The km data in this dataset is separated by a special token \u200b. Loading this dictionary in the Jieba tokenizer, we get a Khmer tokenizer. We preprocess the Khmer data with our Khmer tokenizer followed by BPE subword. We fine tune the three pre-trained models on all data with different sampling strategies. Each pre-trained model is fine tuned with three different sampling strategies and we get 9 models. We use all provided Khmer monolingual data.  The English monolingual data has 90M sentences containing all Newscrawl 2019 data and others sampled from Newscrawl 2014 to 2018. We divide the data into 9 groups. The detailed experiment results are shown in Table~\ref{tab:en_km}. 

\paragraph{Final Submission}
For Khmer$\rightarrow$English, our final submission is an ensemble model consisting of all 9 models after the iterative back-transkation stage. For English$\rightarrow$Khmer, we find the ensemble model has lower BLEU score than the best single model, so we use the best single model as our final submission. The final submissions achieve 51.8 BLEU on wmt20 testset for English$\rightarrow$Khmer direction and 17.6 BLEU for Khmer$\rightarrow$English direction.

\section{Conclusion}
\label{sec:conclusion}

This paper describes VolcTrans's NMT systems for the WMT20 shared news translation task. For all directions, we almost adopted the same strategies, except for low-resource language pairs, we employed multilingual pre-training to boost the baseline models. We found that splitting the monolingual data into disjoint parts is an effective way to increase data diversity among single models, which is an important premise for building strong ensemble models. Our final systems achieved significant improvements, usually 3 to 5 BLEU scores, over baseline systems by integrating techniques such as tagged back-translation, iterative back-translation, random ensemble, knowledge distillation.

\section*{Acknowledgments}
We thank Jun Cao, Zhuo Zhi, Runxin Xu for their support for filtering data, and Zherui Liu for supporting the computing resources.
We would also like to thank the anonymous reviewers for their valuable comments.

\bibliography{ref}

\begin{thebibliography}{18}
\expandafter\ifx\csname natexlab\endcsname\relax\def\natexlab#1{#1}\fi

\bibitem[{Bhojanapalli et~al.(2020)Bhojanapalli, Yun, Rawat, Reddi, and
  Kumar}]{bhojanapalli2020low}
Srinadh Bhojanapalli, Chulhee Yun, Ankit~Singh Rawat, Sashank~J Reddi, and
  Sanjiv Kumar. 2020.
\newblock Low-rank bottleneck in multi-head attention models.
\newblock \emph{arXiv preprint arXiv:2002.07028}.

\bibitem[{Caswell et~al.(2019)Caswell, Chelba, and
  Grangier}]{caswell2019tagged}
Isaac Caswell, Ciprian Chelba, and David Grangier. 2019.
\newblock Tagged back-translation.
\newblock \emph{arXiv preprint arXiv:1906.06442}.

\bibitem[{Edunov et~al.(2018)Edunov, Ott, Auli, and
  Grangier}]{edunov2018understanding}
Sergey Edunov, Myle Ott, Michael Auli, and David Grangier. 2018.
\newblock Understanding back-translation at scale.
\newblock pages 489--500.

\bibitem[{Hinton et~al.(2012)Hinton, Srivastava, Krizhevsky, Sutskever, and
  Salakhutdinov}]{hinton2012improving}
Geoffrey~E Hinton, Nitish Srivastava, Alex Krizhevsky, Ilya Sutskever, and
  Ruslan~R Salakhutdinov. 2012.
\newblock Improving neural networks by preventing co-adaptation of feature
  detectors.
\newblock \emph{arXiv preprint arXiv:1207.0580}.

\bibitem[{Kingma and Ba(2014)}]{kingma2014adam}
Diederik~P Kingma and Jimmy Ba. 2014.
\newblock Adam: A method for stochastic optimization.
\newblock \emph{arXiv preprint arXiv:1412.6980}.

\bibitem[{Li et~al.(2019)Li, Li, Xu, Lin, Liu, Liu, Wang, Zhang, Xu, Wang,
  Feng, Chen, Liu, Li, Wang, Xiao, and Zhu}]{DBLP:conf/wmt/LiLXLLLWZXWFCLL19}
Bei Li, Yinqiao Li, Chen Xu, Ye~Lin, Jiqiang Liu, Hui Liu, Ziyang Wang, Yuhao
  Zhang, Nuo Xu, Zeyang Wang, Kai Feng, Hexuan Chen, Tengbo Liu, Yanyang Li,
  Qiang Wang, Tong Xiao, and Jingbo Zhu. 2019.
\newblock \href {https://doi.org/10.18653/v1/w19-5325} {The niutrans machine
  translation systems for {WMT19}}.
\newblock In \emph{Proceedings of the Fourth Conference on Machine Translation,
  {WMT} 2019, Florence, Italy, August 1-2, 2019 - Volume 2: Shared Task Papers,
  Day 1}, pages 257--266. Association for Computational Linguistics.

\bibitem[{Lin et~al.(2020)Lin, Pan, Wang, Qiu, Feng, Zhou, and Li}]{lin2020pre}
Zehui Lin, Xiao Pan, Mingxuan Wang, Xipeng Qiu, Jiangtao Feng, Hao Zhou, and
  Lei Li. 2020.
\newblock Pre-training multilingual neural machine translation by leveraging
  alignment information.
\newblock In \emph{Proceedings of the 2020 Conference on Empirical Methods in
  Natural Language Processing (EMNLP)}, pages 2649--2663.

\bibitem[{Liu et~al.(2016)Liu, Utiyama, Finch, and
  Sumita}]{DBLP:conf/naacl/LiuUFS16}
Lemao Liu, Masao Utiyama, Andrew~M. Finch, and Eiichiro Sumita. 2016.
\newblock \href {https://doi.org/10.18653/v1/n16-1046} {Agreement on
  target-bidirectional neural machine translation}.
\newblock In \emph{{NAACL} {HLT} 2016, The 2016 Conference of the North
  American Chapter of the Association for Computational Linguistics: Human
  Language Technologies, San Diego California, USA, June 12-17, 2016}, pages
  411--416. The Association for Computational Linguistics.

\bibitem[{Ott et~al.(2019)Ott, Edunov, Baevski, Fan, Gross, Ng, Grangier, and
  Auli}]{ott2019fairseq}
Myle Ott, Sergey Edunov, Alexei Baevski, Angela Fan, Sam Gross, Nathan Ng,
  David Grangier, and Michael Auli. 2019.
\newblock fairseq: A fast, extensible toolkit for sequence modeling.
\newblock \emph{arXiv preprint arXiv:1904.01038}.

\bibitem[{Ott et~al.(2018)Ott, Edunov, Grangier, and Auli}]{ott2018scalenmt}
Myle Ott, Sergey Edunov, David Grangier, and Michael Auli. 2018.
\newblock Scaling neural machine translation.
\newblock \emph{CoRR}, abs/1806.00187.

\bibitem[{Post(2018)}]{post-2018-call}
Matt Post. 2018.
\newblock \href {https://www.aclweb.org/anthology/W18-6319} {A call for clarity
  in reporting {BLEU} scores}.
\newblock In \emph{Proceedings of the Third Conference on Machine Translation:
  Research Papers}, pages 186--191, Belgium, Brussels. Association for
  Computational Linguistics.

\bibitem[{Sun et~al.(2019)Sun, Jiang, Xiong, He, Wu, and Wang}]{sun2019baidu}
Meng Sun, Bojian Jiang, Hao Xiong, Zhongjun He, Hua Wu, and Haifeng Wang. 2019.
\newblock Baidu neural machine translation systems for wmt19.
\newblock In \emph{Proceedings of the Fourth Conference on Machine Translation
  (Volume 2: Shared Task Papers, Day 1)}, pages 374--381.

\bibitem[{Vaswani et~al.(2017)Vaswani, Shazeer, Parmar, Uszkoreit, Jones,
  Gomez, Kaiser, and Polosukhin}]{vaswani2017attention}
Ashish Vaswani, Noam Shazeer, Niki Parmar, Jakob Uszkoreit, Llion Jones,
  Aidan~N Gomez, {\L}ukasz Kaiser, and Illia Polosukhin. 2017.
\newblock Attention is all you need.
\newblock In \emph{Advances in neural information processing systems}, pages
  5998--6008.

\bibitem[{Wang et~al.(2018)Wang, Gong, Zhu, Xie, and Bian}]{wang2018tencent}
Mingxuan Wang, Li~Gong, Wenhuan Zhu, Jun Xie, and Chao Bian. 2018.
\newblock Tencent neural machine translation systems for wmt18.
\newblock In \emph{Proceedings of the Third Conference on Machine Translation:
  Shared Task Papers}, pages 522--527.

\bibitem[{Wu et~al.(2019)Wu, Fan, Baevski, Dauphin, and Auli}]{wu2019pay}
Felix Wu, Angela Fan, Alexei Baevski, Yann~N Dauphin, and Michael Auli. 2019.
\newblock Pay less attention with lightweight and dynamic convolutions.
\newblock \emph{arXiv: Computation and Language}.

\bibitem[{Xia et~al.(2019)Xia, Tan, Tian, Gao, Chen, Fan, Gong, Leng, Luo, Wang
  et~al.}]{xia2019microsoft}
Yingce Xia, Xu~Tan, Fei Tian, Fei Gao, Weicong Chen, Yang Fan, Linyuan Gong,
  Yichong Leng, Renqian Luo, Yiren Wang, et~al. 2019.
\newblock Microsoft research asia's systems for wmt19.
\newblock \emph{arXiv: Computation and Language}.

\bibitem[{Yang et~al.(2019)Yang, Wang, Zhou, Zhao, Yu, Zhang, and
  Li}]{yang2019towards}
Jiacheng Yang, Mingxuan Wang, Hao Zhou, Chengqi Zhao, Yong Yu, Weinan Zhang,
  and Lei Li. 2019.
\newblock Towards making the most of bert in neural machine translation.
\newblock \emph{arXiv preprint arXiv:1908.05672}.

\bibitem[{Zhang et~al.(2018)Zhang, Liu, Li, Zhou, and Chen}]{zhang2018joint}
Zhirui Zhang, Shujie Liu, Mu~Li, Ming Zhou, and Enhong Chen. 2018.
\newblock Joint training for neural machine translation models with monolingual
  data.
\newblock pages 555--562.

\end{thebibliography}
\bibliographystyle{acl_natbib}

\end{document}